\documentclass[journal]{IEEEtran}
\usepackage{amsmath,amsfonts}
\usepackage{algorithmic}
\usepackage{algorithm}
\usepackage{array}
\usepackage[caption=false,font=normalsize,labelfont=sf,textfont=sf]{subfig}
\usepackage{textcomp}
\usepackage{stfloats}
\usepackage{url}
\usepackage{verbatim}
\usepackage{graphicx}
\graphicspath{{pics/}}
\usepackage{cite}
\usepackage{orcidlink}
\begin{document}

\title{Forecasting Ionospheric Irregularities on GNSS Lines of Sight Using Dynamic Graphs with Ephemeris Conditioning}

\author{Mert Can Turkmen \orcidlink{0009-0003-6195-1715}, 
Eng Leong Tan \orcidlink{0000-0003-3003-6007},
and~Yee Hui Lee \orcidlink{0000-0001-6452-9606}
\thanks{Accepted for publication in IEEE Transactions on Geoscience and Remote Sensing, DOI: 10.1109/TGRS.2026.3745099. \copyright~2026 IEEE. Personal use of this material is permitted. Permission from IEEE must be obtained for all other uses, in any current or future media, including reprinting/republishing this material for advertising or promotional purposes, creating new collective works, for resale or redistribution to servers or lists, or reuse of any copyrighted component of this work in other works.}
\thanks{M. C. Turkmen, E. L. Tan, and Y. H. Lee are with the School of
Electrical and Electronic Engineering, Nanyang Technological University
(NTU), Singapore 639798 (e-mail: mertcan001@e.ntu.edu.sg; eeltan@ntu.edu.sg;
eyhlee@ntu.edu.sg).
Corresponding author: Y. H. Lee (e-mail: eyhlee@ntu.edu.sg).}
}

\markboth{}{}

\maketitle

\begin{abstract}
Most data-driven ionospheric models operate on gridded products, which do not preserve the time-varying sampling structure of satellite-based sensing. We instead model the ionosphere as a dynamic graph over ionospheric pierce points, with connectivity that evolves as satellite positions change. Because satellite trajectories are predictable, the graph topology over the forecast horizon can be constructed in advance. We exploit this property to condition forecasts on the future graph structure, which we term \textit{ephemeris conditioning}. This enables prediction on lines of sight (LoS) that appear only in the forecast horizon. We evaluate our framework on Global Navigation Satellite System data from a co-located receiver pair in Singapore spanning 2023 to 2025. The task is forecasting irregularities defined by the Rate of TEC Index (ROTI) up to 2 hours ahead as per-node binary classification. The resulting model, IonoDGNN, achieves a Brier Skill Score (BSS) of 0.55 and an area under the precision--recall curve (PR-AUC) of 0.77. These correspond to improvements over persistence of 53\% in BSS and 58\% in PR-AUC, with larger gains at longer lead times. Ablations confirm that graph structure and ephemeris conditioning each contribute meaningfully. Under simulated coverage dropout, the model retains predictive skill on affected nodes through spatial message passing from observed neighbors. Compared to interpolation baselines, the proposed model achieves better recovery, especially at higher dropout levels. These results suggest that dynamic graph forecasting on evolving LoS is a viable alternative for ionospheric modeling. The project and the dataset are available at \url{https://github.com/Mert-chan/IonoDGNN}.
\end{abstract}


\section{Introduction}
\label{sec:intro}

\IEEEPARstart{G}{lobal} Navigation Satellite System (GNSS) signals propagate through the ionosphere along satellite--receiver paths, with delays proportional to the integrated electron density, or total electron content (TEC)~\cite{Pi1997}. Most data-driven approaches to forecasting TEC~\cite{liu2022ml,yu2024graph,turkmen2025ionobench} rely on gridded ionospheric products~\cite{HernandezPajares2011,Mar23recentGIMguardian}. These products provide stable representations of large-scale ionospheric behavior, but they are not direct measurements. Their estimates depend on model assumptions, how sparse observations are turned into grid values, and availability of input observations~\cite{Chen2020GIMAccuracy,Li2023GIMConsistency}. As a result, they can carry method-dependent biases, with these effects most pronounced at low latitudes and in regions with sparse observation coverage~\cite{Ren2016productbiases,Jer23bRIMs}.

Such regions also exhibit rapid plasma density variations, forming sharp and localized structures such as equatorial plasma bubbles (EPBs)~\cite{Bhattacharyya2022EPB}. These irregularities disrupt radio signals through scattering and scintillation~\cite{Kintner2007}, leading to loss of lock and degraded positioning performance~\cite{Aguiar2025}. Forecasting such behavior is a priority for GNSS applications, but it is also challenging. These structures evolve quickly and on small spatial scales. As a result, the gridding, smoothing, and time-averaging used in map data can weaken or distort the relevant features.

Additionally, models trained on gridded products risk learning the artifacts of the mapping process alongside the physical signal~\cite{Geirhos2020}. The same concern cuts across the existing formulations for irregularity forecasting. Regional-map approaches predict occurrence over spatial grids~\cite{Zhao2025EPB,Atabati2024ConvGRUScintillation,Liu2024HybridEPB}, providing coverage but feeding the model targets shaped by approximation. There are also single-station approaches that forecast proxy indices such as Rate of TEC Index (ROTI) or amplitude scintillation index ($S_4$) at individual receivers~\cite{Atabati2021,Trachuentong2024Scintillation,Carvalho2022ScintillationNowcasting}. These capture local dynamics but do not produce a regional picture.

Recent studies have introduced graph neural networks (GNNs) to ionospheric modeling. However, they operate on fixed grid nodes~\cite{Chen2025Atmosphere,yu2024graph,Kelebek2025} or static station graphs~\cite{Kaselimi2025}, leaving the underlying representation unchanged.

This motivates a different viewpoint. The ionosphere is not observed on a fixed grid but through satellite--receiver lines of sight (LoS) whose sampling structure evolves as satellites move. At any instant, GNSS observes the ionosphere at a set of ionospheric pierce points (IPPs) that move, appear, and disappear as satellites rise and set. This evolving LoS structure shapes what is observed, where, and with what spatial coverage. Prior work has modeled ionospheric quantities along individual LoS~\cite{Cai2024,Mao2024}, but not the evolving spatial structure among them. Dynamic GNNs (DGNNs), where the node and edge sets change at each timestep, offer a natural formalism for this structure. They have also shown promise in domains with similarly transient sensor configurations~\cite{Xu2023GDGCN,Tang2025DAHG}. We apply this perspective to GNSS in this study.

Our model, IonoDGNN (Ionospheric DGNN), operates directly on GNSS LoS observations where each node is an IPP carrying observation and ephemeris-derived features, and edges encode spatial relationships among simultaneous IPPs. Missing observations are simply absent rather than imputed, and the model adapts to the transient observation density at each timestep. It therefore learns from only the observed signal without inheriting biases from mapping approximations.

Our key novelty is what we term \emph{ephemeris conditioning}, which leverages the near-deterministic nature of satellite orbits. Since orbit predictions are available up to 24 hours ahead with sub-meter accuracy~\cite{IGS_IGSUSUM_2000}, the principle of conditioning on known future scalars~\cite{lim2021TFT} can be extended to the future GNSS observation graphs. The model can therefore reason about where the ionosphere will be sampled over the forecast window. This is critical since new satellites regularly enter the observation set within the prediction window. These introduce LoS with no observed history, which we refer to as \emph{no-history nodes}.

As a proof of concept, we apply the proposed approach to forecasting ROTI-defined irregularities at 5-minute cadence up to 2 hours ahead as binary per-node event prediction from multi-GNSS observations. The data configuration uses a zero-baseline receiver pair in Southeast Asia. We take advantage of this co-location to cross-validate detected events, yielding consistent labels for training and evaluation under realistic uncertainty.

Because irregularity occurrence is heavily imbalanced toward quiet-time samples, aggregate metrics such as RMSE, R$^2$, or accuracy can under-represent performance on the rare event class~\cite{Atabati2021,Liu2024HybridEPB,Trachuentong2024Scintillation}. Event-focused, probabilistic, and contingency-based verification have been used in some studies~\cite{Chartier2018ScintillationML,Carter2020EPBAssessment,Zhao2025EPB}, and we build on this direction with a broader metric set aimed at event detection skill. Our evaluation further includes ablations that isolate the contributions of graph representation and ephemeris conditioning, with a persistence baseline as a reference for comparison. In addition, a coverage-dropout analysis examines prediction on nodes with missing observations and compares the model's ability to recover these nodes against spatial interpolation baselines.

In summary, this paper makes two main contributions:
\begin{enumerate}
    \item We introduce IonoDGNN, a dynamic graph formulation for ionospheric modeling that represents GNSS LoS observations as evolving graph nodes without intermediate gridding.
    \item We propose ephemeris conditioning, enabling better overall performance and generalization to no-history nodes by leveraging the future graph topology constructed from satellite ephemeris.
\end{enumerate}

\section{Data and Preprocessing}

\subsection{Observation Network and Period}

The dataset uses multi-GNSS observations~\cite{Dow2009IGS} from a receiver pair in Singapore: NTUS and SIN1. Both stations track five constellations (GPS, Galileo, BeiDou, QZSS, GLONASS) at 30-second sampling. The dataset spans March 2023 through April 2025, corresponding to a high-activity portion of Solar Cycle~25 and thus capturing more frequent ionospheric disturbances including EPBs and related scintillation~\cite{Tongkasem2025IIRSolarCycle25}.

Precise satellite orbits are derived from the Wuhan University ultra-rapid combined products~\cite{Li2022iWUH}. Hourly Dst and daily F10.7 indices are sourced from NASA's OMNIWeb database~\cite{King2005}.

\subsection{Feature Derivation}

Per-satellite ROTI is derived from dual-frequency geometry-free carrier phase following the original definition of Pi~et~al.~\cite{Pi1997}. The geometry-free combination isolates the ionospheric delay. From this delay, relative slant total electron content (TEC), rate of TEC (ROT), and its 5-minute standard deviation, the ROT index (ROTI), are computed with a 30-second cadence.

At each processing step, we apply a $30^{\circ}$ elevation mask, a 25\,dB-Hz signal-to-noise ratio (SNR) cutoff, and artifact detection via geometry-free phase jumps adapted from~\cite{Li2022gfphasethresh}. IPP coordinates are computed on a 450\,km thin-shell model and converted to quasi-dipole magnetic coordinates via ApexPy~\cite{Emmert2010}. All computed parameters are downsampled to 5-minute cadence for modeling.

\begin{figure*}[!t]
\centering
\includegraphics[width=\textwidth]{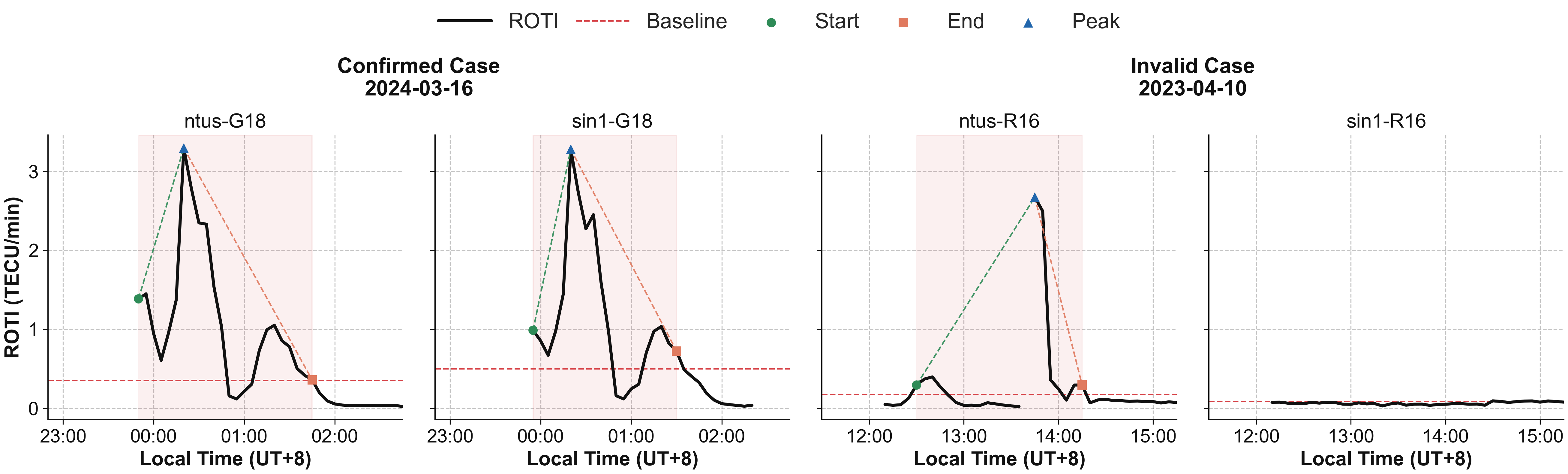}
\caption{Examples of cross-station QC outcomes. (a)~Confirmed event: both NTUS and SIN1 detect a coincident ROTI enhancement on the same satellite. (b)~Invalid segment: only one station registers elevated ROTI while the other remains quiet.}
\label{fig:fig_qc_examples}
\end{figure*}

\subsection{Quality Control (QC) Labels}
\label{sec:qc_labels}

We currently lack multiple instruments to independently confirm ionospheric disturbances. However, the two selected stations form a zero-baseline pair ($\sim$400\,m separation), which we exploit for quality control. Since both receivers sample effectively the same ionospheric volume at L-band Fresnel scales~\cite{Aguiar2025}, a physical irregularity must appear at both.

Per-satellite ROTI irregularities are segmented in each time series using a heuristic-based event detector that identifies contiguous intervals of elevated activity and determines the peak amplitude and the start and end of the event. Segments whose peak ROTI falls below 0.3\,TECU/min are reclassified as quiet. This value separates the bell-shaped quiet-time mode from the long-tailed event distribution on a log-scaled amplitude histogram. For segments whose peak exceeds this threshold, the peak is traced forward and backward to the nearest baseline condition, and all timesteps within this interval are labeled as event.

Each detected event is then cross-validated against the second station and assigned one of four labels:

\begin{itemize}
    \item \textbf{Confirmed (+1):} Both stations detect an event on the same satellite, and the temporal intersection of the two segments covers at least 50\% of the shorter segment.
    \item \textbf{Quiet (0):} No event detected.
    \item \textbf{Unverified ($-$1):} The other station lacks data coverage during the event window; cross-validation is not possible.
    \item \textbf{Invalid ($-$2):} The other station has valid, quiet data during the segment; the event is likely a receiver-side artifact.
\end{itemize}
Confirmation is a segment-level decision: once a segment is corroborated, every timestep within it inherits the event label, without requiring second-station coverage at each epoch. Fig.~\ref{fig:fig_qc_examples} shows examples of confirmed and invalid segments. This procedure yields confirmed events and quiet intervals for training, while also flagging ambiguous cases (unverified and invalid) that are excluded from training but retained for further analysis (Sec.~\ref{sec:ambiguous}).

\begin{table*}[!t]
\centering
\caption{Dataset Statistics}
\label{tab:datasplit_stats}
\begin{tabular}{lrrrrrr}
\hline
Split & Timesteps & Nodes & Event (\%) & Quiet (\%) & Unverified (\%) & Invalid (\%) \\
\hline
Train & 120{,}471 & 3{,}097{,}829 & 266{,}022 (8.6) & 2{,}764{,}013 (89.2) & 65{,}773 (2.1) & 2{,}021 (0.1) \\
Val   &  22{,}280 &   570{,}575 &  59{,}911 (10.5) &   497{,}118 (87.1) & 13{,}207 (2.3) &   339 (0.1) \\
Test  &  30{,}369 &   797{,}339 &  53{,}590 (6.7) &   706{,}190 (88.6) & 34{,}243 (4.3) &   3{,}316 (0.4) \\
\hline
\end{tabular}
\end{table*}

\subsection{Data Split}
\label{sec:data_split}

The dataset period is selected based on the simultaneous availability of both stations and then split chronologically. Training spans 17 March 2023 20:50 UTC to 31 August 2024 23:55 UTC, validation 1 September 2024 02:00 UTC to 16 December 2024 01:00 UTC, and test 16 December 2024 03:05 UTC to 1 April 2025 00:00 UTC. Consecutive subsets are separated by gaps of more than 2 hours to prevent information leakage between subsets. In addition, split boundaries are placed so that each subset remains representative of the data distribution without breaking chronological order. Timesteps containing only unverified or invalid segments, with no confirmed events on any satellite, are excluded from training and validation to control training noise. When confirmed and ambiguous labels coexist at the same timestep across different satellites, the timestep is retained, but unverified and invalid IPPs are masked from the loss function (see Section~\ref{sec:training}).

As shown in Table~\ref{tab:datasplit_stats}, the dataset maintains high data availability across all splits, totaling more than 173{,}000 timesteps with 4.5 million multi-GNSS observations. Across all timesteps, the observed event rate is $\approx$7\%, which we take as the climatological rate for metric calculation (Section~\ref{sec:evaluation}).

\begin{figure*}[!t]
\centering
\includegraphics[width=\textwidth]{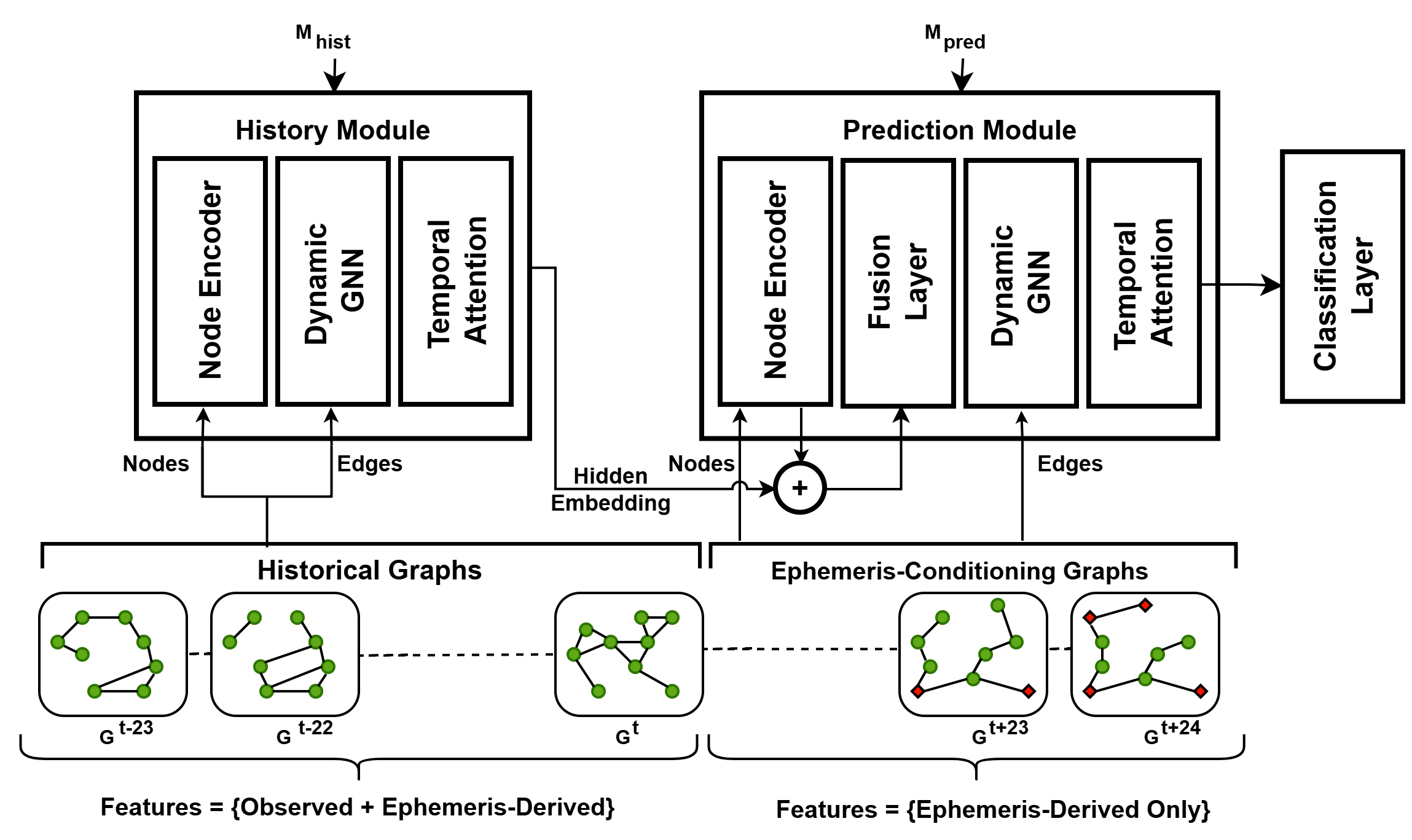}
\caption{IonoDGNN overview. (a) Dynamic observation graphs over the history and forecast windows, with observed, future, and no-history nodes. (b) History module: at each history timestep, a node encoder followed by spatial GNN layers processes the graph, and temporal attention aggregates the sequence into the history embedding $\mathbf{h}_{\mathrm{hist}}$. (c) Prediction module: at each forecast step, $\mathbf{h}_{\mathrm{hist}}$ is combined with the encoded ephemeris features through a fusion MLP before the spatial GNN layers; temporal attention over the forecast horizon feeds a classification head that outputs per-node event probabilities.}
\label{fig:architecture}
\end{figure*}

\section{Methodology}
\label{sec:method}
This section describes the two core contributions: the dynamic observation graph (Section~\ref{sec:graph_construction}) and ephemeris conditioning (Section~\ref{sec:ephemeris_conditioning}). The remaining subsections cover implementation details: variable-size graph handling (Section~\ref{sec:alignment}), masks (Section~\ref{sec:masks}), model architecture (Section~\ref{sec:architecture}), and training and evaluation (Sections~\ref{sec:training}--\ref{sec:evaluation}).

\subsection{Dynamic Observation Graph}
\label{sec:graph_construction}

At each 5-minute timestep $t$, we construct a graph
$G^t = (\mathcal{V}^t, \mathcal{E}^t, \mathbf{X}_V^t, \mathbf{X}_E^t)$ from the set of active IPPs, where $\mathcal{V}^t$ and $\mathcal{E}^t$ are the node and edge sets, and $\mathbf{X}_V^t \in \mathbb{R}^{|\mathcal{V}^t| \times d_V}$ and $\mathbf{X}_E^t \in \mathbb{R}^{|\mathcal{E}^t| \times d_E}$ are the node and edge feature matrices. Each node corresponds to a visible satellite--receiver LoS link at time $t$. As satellites rise and set, both $\mathcal{V}^t$ and $\mathcal{E}^t$ vary over time. We therefore treat the task as node-and-edge-varying discrete-time dynamic graph forecasting, following the taxonomy of~\cite{yang2024dgnnsurvey}, with history and forecast windows of 2 hours (Fig.~\ref{fig:architecture}(a)).

Edges connect each node to its $K$-nearest neighbors ($K$-NN) by haversine distance on the IPP shell. Since the number of active LoS varies with satellite visibility, each node is connected to $\min(K, |\mathcal{V}^t| - 1)$ neighbors, so a snapshot with $|\mathcal{V}^t| \le K$ reduces to a fully connected graph. No padded or synthetic neighbors are introduced. Because the edge features include geodetic distance and bias the attention scores (Section~\ref{sec:architecture}), the model can still attenuate spatially distant neighbors in such snapshots. At each snapshot, the edges carry the features $\mathbf{X}_E^t$: geodetic distance, differences in geographic and geomagnetic coordinates, angular separation in local solar time, and a binary flag indicating whether two nodes share the same satellite across the zero-baseline pair.

The node feature matrix splits column-wise into an observed and an ephemeris-derived block, $\mathbf{X}_V^t = [\,\mathbf{X}_{\mathrm{obs}}^t \,;\, \mathbf{X}_{\mathrm{eph}}^t\,]$, with the following features:
\begin{itemize}
    \item Observed block ($\mathbf{X}_{\mathrm{obs}}^t$): ROTI, TEC rate ($\Delta \mathrm{TEC} / \Delta t$), effective SNR (minimum of the signal pair), storm-time index (Dst), solar flux index (F10.7). Available only in the history window ($\tau \le t$).
    \item Ephemeris-derived block ($\mathbf{X}_{\mathrm{eph}}^t$): IPP latitude/longitude (geographic and quasi-dipole magnetic), elevation angle, cyclical encodings of day of year and local solar time. Computable from ephemerides and time information, hence available in both windows; future graphs carry this block only, $\mathbf{X}_V^{\tau} = [\,\mathbf{X}_{\mathrm{eph}}^{\tau}\,]$ for $\tau > t$.
\end{itemize}

All features are min--max normalized using training-period statistics.

\subsection{Handling Varying Nodes and Edges}
\label{sec:alignment}

Node identities (station, constellation, satellite) are known and consistent across time. For each sample, we construct a unified node index $\mathcal{V}_{\cup} = \bigcup_{\tau=t-23}^{t+24} \mathcal{V}^{\tau}$ spanning both the history and prediction windows. Every identity in $\mathcal{V}_{\cup}$ is assigned a fixed position $v \in \{1,\ldots,N_{\cup}\}$, where $N_{\cup} = |\mathcal{V}_{\cup}|$, into which per-timestep features and labels are written. Edge indices at each timestep are remapped accordingly.

Across a batch of $B$ samples, $N_{\cup}$ varies. We pad all tensors to $N_{\max} = \max_{b} N_{\cup}^{(b)}$, yielding shape $[B,\, 48,\, N_{\max}]$. Following the disjoint-graph mini-batch strategy of~\cite{Fey2019PyG}, message passing runs over the full batch by expressing edge indices in a flattened space of size $B \cdot N_{\max}$, with per-sample offsets of $b \cdot N_{\max}$ to keep graphs disconnected.

Node identities serve only for alignment, padding, and mask construction; they are never provided as input features. This batching procedure is tailored to our setting where per-timestep graphs are small (mean $\approx 26$ nodes), so padding overhead is negligible.

\subsection{Masks}
\label{sec:masks}

Four binary masks of shape $\{0,1\}^{B \times 24 \times N_{\max}}$ are defined over the padded unified node axis to control what the model sees, how it learns, and what it predicts on. $\mathbf{M}_{\mathrm{hist}}$ is indexed over the history timesteps, while $\mathbf{M}_{\mathrm{pred}}$, $\mathbf{M}_{\mathrm{valid}}$, and $\mathbf{M}_{\mathrm{nohist}}$ are indexed over the forecast timesteps.

$\mathbf{M}_{\mathrm{hist}}(b,\tau,v) = 1$ indicates that node $v$ exists at history timestep $\tau$ for sample $b$. It gates history feature encoding and history-time attention, ensuring that only input graph nodes contribute to the history context embedding.

$\mathbf{M}_{\mathrm{pred}}$ plays the analogous role in the forecast window: it gates ephemeris feature encoding and forecast-time attention. Fig.~\ref{fig:architecture}(a) illustrates how $\mathbf{M}_{\mathrm{hist}}$ and $\mathbf{M}_{\mathrm{pred}}$ gate the historical and future graph windows.

$\mathbf{M}_{\mathrm{valid}}$ marks nodes with confirmed labels from the quality-control procedure (Section~\ref{sec:qc_labels}). Unverified and invalid nodes receive predictions at inference but do not contribute gradients during training.

$\mathbf{M}_{\mathrm{nohist}}$ flags nodes that appear only in the prediction window, having no presence in the history graphs: neither the node nor its edges exist in the history window. These are the no-history nodes defined in Section~\ref{sec:intro}. This mask is used for evaluating performance on no-history nodes, whose missing history embeddings are handled as described in Section~\ref{sec:ephemeris_conditioning}.

\subsection{Ephemeris Conditioning}
\label{sec:ephemeris_conditioning}

For each prediction step $t{+}k$, ephemerides yield IPP locations, elevation angles, and local solar time for every active satellite--receiver link. From these, we compute the ephemeris-derived features $\mathbf{x}_{\mathrm{eph},v}^{t+k}$ (a row of $\mathbf{X}_{\mathrm{eph}}^{t+k}$) and the $K$-NN edge set $\mathcal{E}^{t+k}$ exactly as in the history window (Section~\ref{sec:graph_construction}).

Satellites that enter the elevation filter during the forecast horizon introduce no-history nodes. Since no identity information is used as input, the model cannot draw on satellite-specific representations. Without ephemeris conditioning, such nodes carry no input information and collapse to the learned output bias (Section~\ref{sec:overall_performance}).

Ephemeris conditioning addresses this by providing $\mathbf{x}_{\mathrm{eph},v}^{t+k}$ and a well-defined neighborhood in $\mathcal{E}^{t+k}$ for every forecast node. As illustrated in Fig.~\ref{fig:architecture}(c), the per-node history embedding $\mathbf{h}_{\mathrm{hist}}$ produced by the history module is combined with the corresponding forecast-step node encoding through a fusion MLP. Nodes flagged by $\mathbf{M}_{\mathrm{nohist}}$ are excluded from history encoding by the history mask $\mathbf{M}_{\mathrm{hist}}$, so their $\mathbf{h}_{\mathrm{hist}}$ reduces to a learned constant that carries no observational information. Their context is instead recovered through spatial message passing from neighboring IPPs that carry historical state.

\subsection{Model Architecture}
\label{sec:architecture}

IonoDGNN separates input processing into a history module and a prediction module (Fig.~\ref{fig:architecture}(b)--(c)):

\begin{itemize}
    \item \textbf{History module:} At each history step $\tau = t{-}23,\ldots,t$, a shared node encoder maps the node features $\mathbf{X}_V^{\tau} = [\mathbf{X}_{\mathrm{obs}}^{\tau};\mathbf{X}_{\mathrm{eph}}^{\tau}]$ to node embeddings. A spatial GNN, implemented as multi-head graph attention~\cite{Brody2021gat} with the edge features $\mathbf{X}_E^{\tau}$ biasing the attention scores, then propagates information across that step's graph. Temporal self-attention~\cite{Vaswani2017attention} with sinusoidal positional encoding, masked by $\mathbf{M}_{\mathrm{hist}}$, links the per-step embeddings of each node. The resulting per-node summary $\mathbf{h}_{\mathrm{hist}}$ is then passed to the prediction module.
    \item \textbf{Prediction module:} The prediction module operates on the future graphs $G^{t+1},\ldots,G^{t+24}$ with ephemeris-only features $\mathbf{X}_V^{\tau} = [\mathbf{X}_{\mathrm{eph}}^{\tau}]$. At each forecast step, a node encoder embeds these features, a fusion MLP combines the embedding with $\mathbf{h}_{\mathrm{hist}}$, and a spatial GNN propagates information across the predicted graph. A second temporal attention, masked by $\mathbf{M}_{\mathrm{pred}}$, couples the embeddings across the forecast horizon, and a classification head yields the per-node, per-step disturbance probabilities $\mathbf{P}^{\,t+1:t+24}$.
\end{itemize}

\subsection{Training}
\label{sec:training}

The model minimizes cross-entropy over the prediction window, masked by $\mathbf{M}_{\mathrm{valid}}$. The checkpoint with the highest validation Brier Skill Score (BSS) is selected for testing. Training uses a cosine-annealing scheduler with warmup and the AdamW optimizer. At a fixed batch size of 64, the learning rate was selected via grid search and the neighbor count via a sweep over $K$; performance degraded beyond $K = 14$. Full hyperparameter configurations are available in the model repository at \url{https://github.com/Mert-chan/IonoDGNN/tree/master/source/exp_runs/ablation}.

\subsection{Evaluation}
\label{sec:evaluation}

Our primary metric is the Brier Skill Score $\mathrm{BSS} = 1 - \mathrm{BS}/\mathrm{BS}_{\mathrm{clim}}$. Here, $\mathrm{BS} = \frac{1}{n}\sum_{j=1}^{n} (p_j - y_j)^2$ is the Brier Score over all $n$ scored node--lead-time pairs $(p_j, y_j)$, with $p_j$ the predicted disturbance probability from $\mathbf{P}^{\,t+1:t+24}$ (Section~\ref{sec:architecture}) and $y_j \in \{0,1\}$ the corresponding label (Section~\ref{sec:qc_labels}). The reference $\mathrm{BS}_{\mathrm{clim}} = \bar{y}(1-\bar{y})$ is the climatological Brier Score, with $\bar{y} \approx 0.07$ the climatological event rate (Section~\ref{sec:data_split}). $\mathrm{BSS} = 0$ indicates no skill beyond climatology; $\mathrm{BSS} = 1$ indicates perfect forecasts.

We additionally report area under the receiver operating characteristic curve (ROC-AUC) and area under the precision--recall curve (PR-AUC) as threshold-independent discrimination measures. PR-AUC is particularly informative under class imbalance~\cite{saito2015precision}. Lastly, we report the probability of detection (POD), false alarm ratio (FAR), and critical success index (CSI) at the CSI-optimal decision threshold of each model. All metrics are computed over the full 24-step (2-hour) forecast window.

We compare the full model against three ablations and a persistence baseline (see Fig.~\ref{fig:architecture}(b)--(c) for the components), where ``w/o'' denotes ``without'':
\begin{itemize}
    \item \textbf{IonoDGNN (full):} history module, graph message passing, and prediction module.
    \item \textbf{w/o conditioning:} history module only; predictions from the history embedding alone, no future graphs.
    \item \textbf{w/o graph:} graph message passing removed; nodes processed with node encoder and temporal attention only.
    \item \textbf{w/o history:} prediction module only, with ephemeris features and future graph topology.
    \item \textbf{Persistence:} last observed label as the forecast; no-history nodes receive the climatological event rate.
\end{itemize}
For no-history nodes, we additionally compare against two spatial interpolation baselines, ordinary kriging and inverse distance weighting (IDW), further explained in Section~\ref{sec:nohistory}.

\section{Results}

\subsection{Overall Performance}
\label{sec:overall_performance}

\begin{table*}[!t]
\centering
\caption{Overall Forecasting Performance on All Test Nodes}
\label{tab:overallperformance}
\begin{tabular}{l cccccc}
\hline
Model & BSS$\uparrow$ & ROC-AUC$\uparrow$ & PR-AUC$\uparrow$ & POD$\uparrow$ & FAR$\downarrow$ & CSI$\uparrow$ \\
\hline
IonoDGNN (full)  & \textbf{0.552} & \textbf{0.971} & \textbf{0.773} & \textbf{0.727} & 0.332 & \textbf{0.534} \\
w/o conditioning & 0.489 & 0.944 & 0.701 & 0.643 & 0.338 & 0.484 \\
w/o graph        & 0.401 & 0.951 & 0.650 & 0.654 & 0.460 & 0.420 \\
Persistence      & 0.360 & 0.833 & 0.490 & 0.609 & \textbf{0.276} & 0.494 \\
w/o history      & 0.194 & 0.921 & 0.433 & 0.665 & 0.631 & 0.311 \\
\hline
\end{tabular}
\end{table*}

Table~\ref{tab:overallperformance} summarizes forecasting performance on all test nodes across all lead times. IonoDGNN achieves the highest skill (BSS 0.552, ROC-AUC 0.971, PR-AUC 0.773). The full model detects 73\% of event nodes (POD 0.727) with a CSI of 0.534, outperforming all ablations.

Removing ephemeris conditioning (w/o conditioning) lowers BSS from 0.552 to 0.489 and PR-AUC from 0.773 to 0.701. Removing graph message passing (w/o graph) reduces BSS to 0.401, confirming that spatial context through the $K$-NN topology contributes meaningfully to event discrimination. Removing the history module, and hence the observations entirely (w/o history), yields the weakest overall skill (BSS 0.194, PR-AUC 0.433) but still shows skill above climatology. This demonstrates that future graph topology and ephemeris-derived features alone carry predictive information about irregularity occurrence. The persistence baseline achieves the lowest FAR (0.276) by predicting events only where one is already occurring, but its PR-AUC (0.490) falls below all learned variants except w/o history.

\begin{figure*}[!tb]
\centering
\includegraphics[width=\textwidth]{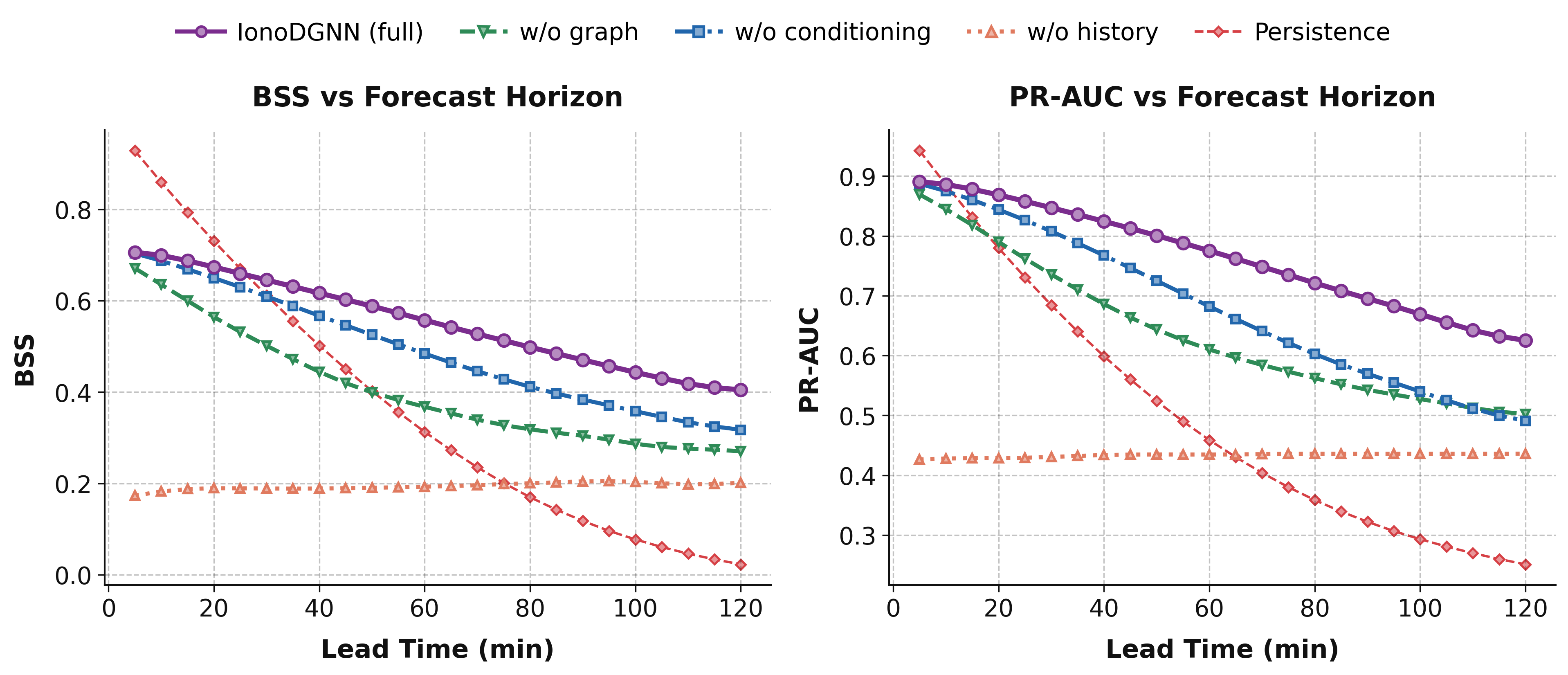}
\caption{Forecast skill as a function of lead time for all model variants. (a)~Brier Skill Score. (b)~Precision--recall AUC.}
\label{fig_lead_time}
\end{figure*}

\subsection{Lead-Time Degradation}
\label{sec:lead_time}

Fig.~\ref{fig_lead_time} shows BSS and PR-AUC as a function of forecast lead time. IonoDGNN degrades most gracefully among models with access to history: at 120 minutes it retains BSS $\approx$ 0.40 and PR-AUC $\approx$ 0.62, while persistence falls to BSS $\approx$ 0.03 and PR-AUC $\approx$ 0.25, approaching climatological skill.

At short lead times ($<$20 min), persistence is very hard to beat, reflecting the strong temporal autocorrelation of the ionosphere~\cite{Muhtarov1999Autocorrelation}. IonoDGNN overtakes it by roughly 25 minutes in BSS and earlier ($\sim$10 min) in PR-AUC, and the gap widens to more than an order of magnitude in BSS by the 2-hour horizon. The w/o conditioning variant tracks IonoDGNN at the shortest horizons but diverges beyond roughly 20--25 minutes. The w/o graph variant degrades faster and remains below both variants across most of the forecast window. This indicates that spatial message passing provides additive benefit at all lead times.

The w/o history variant is qualitatively different: BSS and PR-AUC remain essentially flat across the entire forecast window ($\approx$ 0.20 and $\approx$ 0.43, respectively). This follows naturally, as there is no temporal context to lose without historical observations. It also serves as a sanity check that the masking logic in Section~\ref{sec:masks} introduces no information leakage from observed features into the forecast horizon.

\begin{table*}[!t]
\centering
\caption{Forecasting Performance on No-History Nodes}
\label{tab:nohistory}
\begin{tabular}{l cccccc}
\hline
Method & BSS$\uparrow$ & ROC-AUC$\uparrow$ & PR-AUC$\uparrow$ & POD$\uparrow$ & FAR$\downarrow$ & CSI$\uparrow$ \\
\hline
IonoDGNN (full)  & \textbf{0.529} & \textbf{0.959} & \textbf{0.614} & 0.612 & \textbf{0.424} & \textbf{0.422} \\
Ordinary kriging & 0.432 & 0.953 & 0.550 & 0.574 & 0.488 & 0.371 \\
IDW              & 0.411 & 0.954 & 0.561 & \textbf{0.627} & 0.510 & 0.380 \\
\hline
\end{tabular}
\end{table*}

\subsection{Performance on No-History Nodes}
\label{sec:nohistory}

Ephemeris conditioning allows the model to make predictions for no-history nodes without any observed features. To test whether this capability outperforms spatial interpolation, we compare IonoDGNN against two baselines commonly used in the literature: ordinary kriging and IDW~\cite{Li2018Interpolation}. Both methods interpolate a no-history node from the model predictions at the surrounding history-observed nodes at the same forecast step. Ordinary kriging uses a spherical variogram fitted per forecast snapshot to the source predictions, while IDW uses a power parameter of 2.0 and interpolates from the $\min(K, |\mathcal{V}^{\tau}| - 1)$ nearest source nodes, where $K = 14$ matches the IonoDGNN setting.

Table~\ref{tab:nohistory} shows that IonoDGNN retains most of its overall skill on these nodes (BSS 0.529, ROC-AUC 0.959, PR-AUC 0.614). It outperforms both interpolation baselines on every metric except POD, where IDW detects slightly more events (0.627 vs.\ 0.612) at the cost of a much higher FAR (0.510 vs.\ 0.424). The relative BSS improvements are 22\% over ordinary kriging and 29\% over IDW. This is expected since IonoDGNN can leverage the history embedding of neighboring nodes, which carries information about the recent state of the ionosphere. The interpolation baselines, in contrast, rely solely on the spatial configuration of the future graph. They can therefore only be as good as the model predictions at the future nodes they interpolate from.

\subsection{Event Analysis}
\label{sec:event_vis}

\begin{figure*}[!t]
\centering
\includegraphics[width=\textwidth]{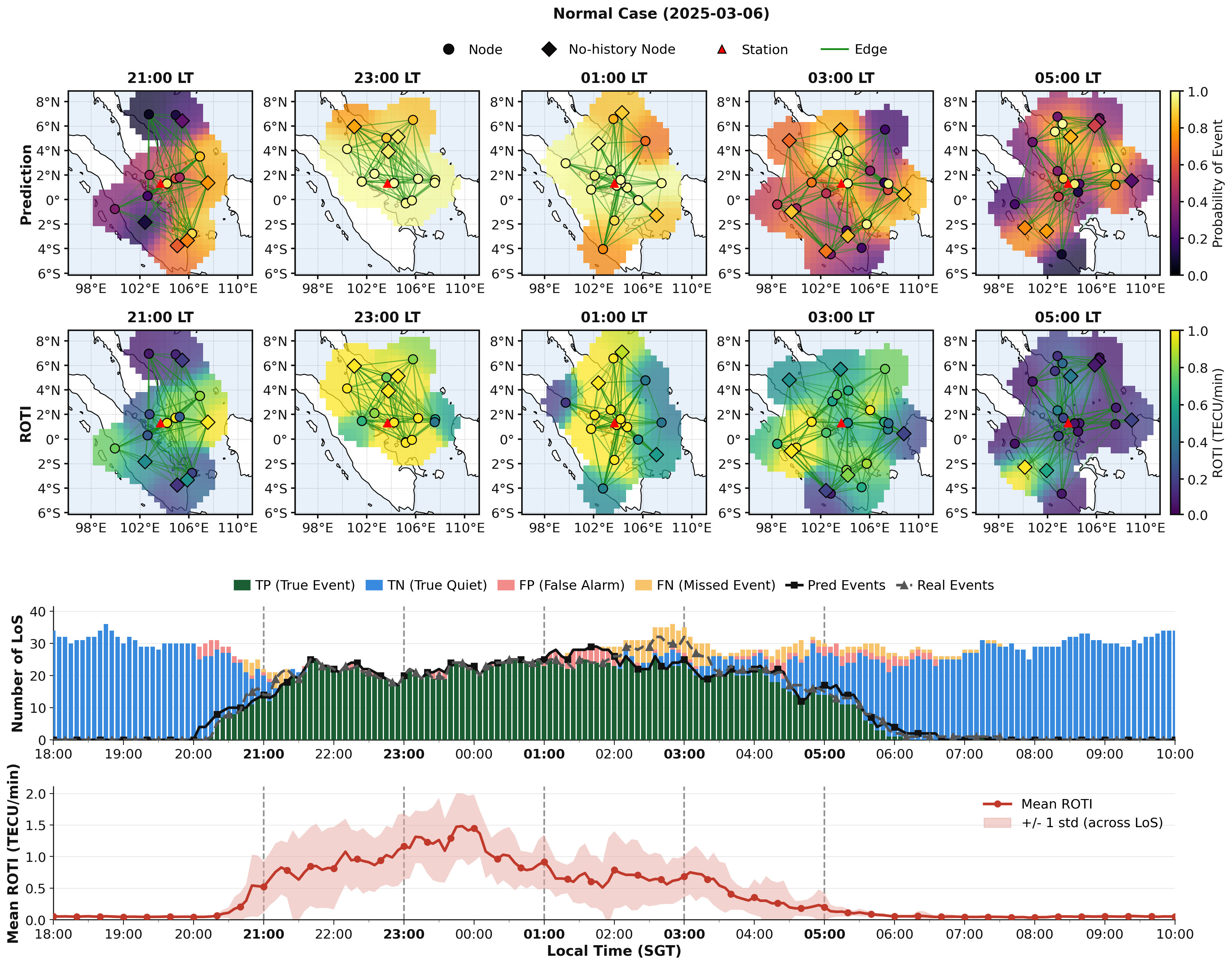}
\caption{Event 1, normal-coverage case (2025-03-06), all predictions at +60-min lead time. First row: predicted event probabilities at five snapshots (circles: nodes; diamonds: no-history nodes; green lines: $K$-NN edges; red triangle: co-located station pair); second row: observed ROTI at the same snapshots; third row: active LoS per timestep decomposed into true positives, true negatives, false alarms, and missed events at the CSI-optimal threshold, with predicted and observed event counts; fourth row: mean ROTI across active LoS with a $\pm 1$ standard deviation band. Dashed lines mark the snapshot times, and background shading interpolates node values for visual context only. The first two rows show different quantities: predicted probability that a node lies inside a labeled event, and the observed ROTI amplitude from which the segment labels are derived (Section~\ref{sec:qc_labels}).}
\label{fig_event1}
\end{figure*}

\begin{figure*}[!t]
\centering
\includegraphics[width=\textwidth]{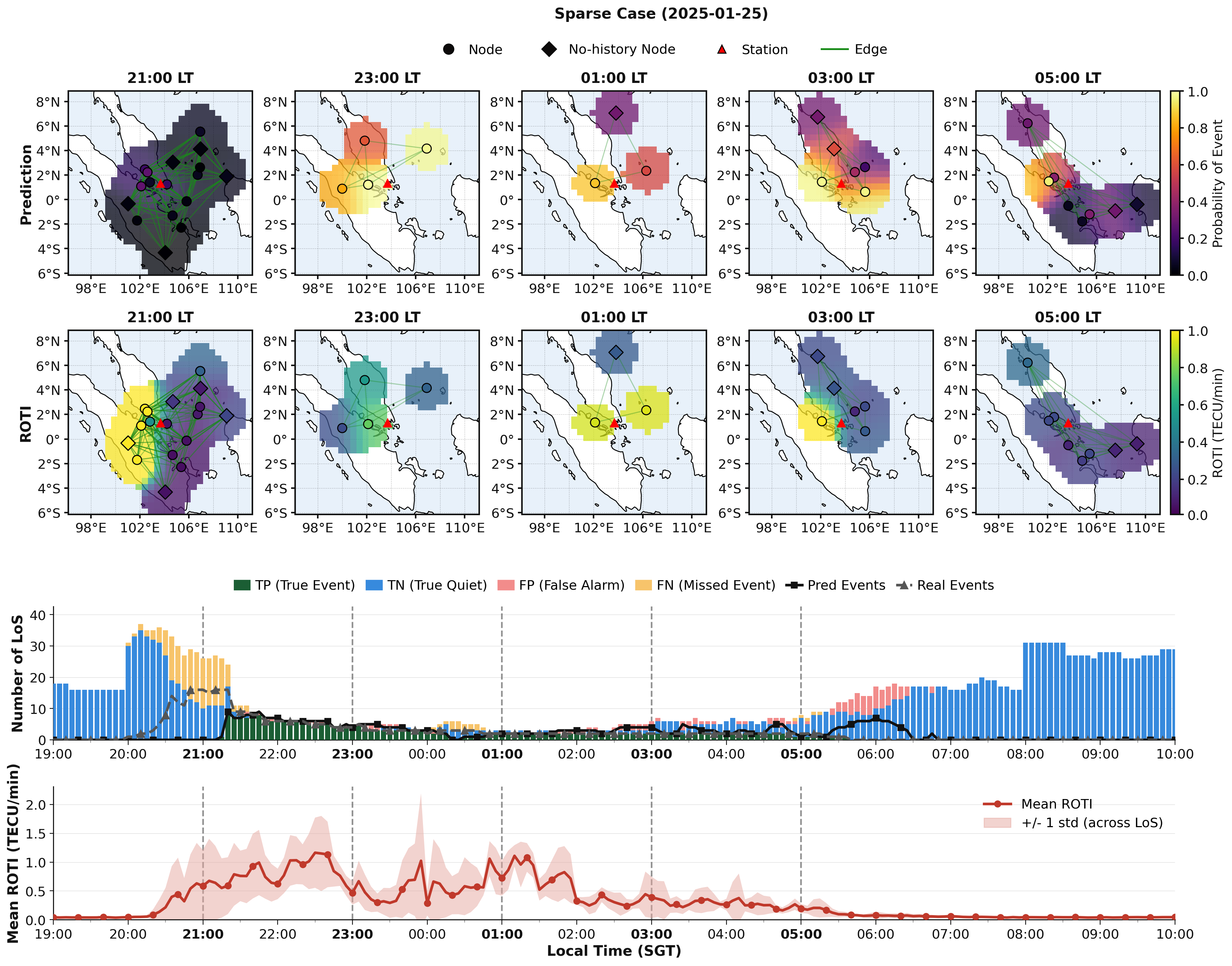}
\caption{Event 2, sparse-coverage case (2025-01-25), in the same layout as Fig.~\ref{fig_event1}. The testable node set collapses during an extended single-receiver data outage.}
\label{fig_event2}
\end{figure*}

To examine how the model captures the spatiotemporal evolution of irregularities, we analyze two contrasting events from the test set at a fixed +60\,min lead time (12 steps): a strong post-sunset event under normal observation coverage (Fig.~\ref{fig_event1}) and an event under sparse coverage (Fig.~\ref{fig_event2}). Both figures share a four-row layout. The first two rows map the predicted event probabilities and the observed ROTI at the same five snapshots spanning the event period. These are different quantities: the model outputs the probability that a node is in the event state, whereas ROTI is the observed amplitude on which the segment labels are based (Section~\ref{sec:qc_labels}). The two rows are therefore compared qualitatively by the spatial pattern of elevated values rather than value by value, since a node with momentarily low ROTI can still carry an event label near the start or end of a segment. The third row summarizes the full night as the number of active LoS per timestep, decomposed into true positives, true negatives, false alarms, and missed events at the CSI-optimal threshold (Section~\ref{sec:evaluation}), together with the predicted and observed event counts. The fourth row shows the mean ROTI across active LoS with a $\pm 1$ standard deviation band, summarizing the overall activity level. Dashed vertical lines mark the five map snapshots.

\textbf{Event 1 (2025-03-06):} Fig.~\ref{fig_event1} presents a strong post-sunset event under normal observation coverage for our setup ($\approx$20--35 active LoS per timestep). The night divides into four phases. In the quiet lead-in (18:00--20:00 LT), predictions remain low across all nodes, with only a brief cluster of false alarms; given the lead time, these reflect the model anticipating the onset at some nodes. During onset and growth (20:15--21:30 LT), the predicted and observed event counts climb together. The 21:00 LT snapshot shows that the model also captures where the event develops: predicted probabilities split between the disturbed east sector ($0$--$1^{\circ}$N, $106$--$108^{\circ}$E) and the still-quiet north ($4$--$8^{\circ}$N), mirroring the observed ROTI gradient. Through the peak phase (21:30--01:00 LT), mean ROTI fluctuates near 1\,TECU/min, peaking at $\approx$1.5\,TECU/min, with individual LoS reaching up to 2\,TECU/min. In this phase, nearly every active node is in the event state, and the model matches this saturation with uniformly high probabilities (23:00 and 01:00 LT snapshots). Errors concentrate in the decay phase (01:00--05:30 LT): as the irregularities become patchy and intermittent (03:00 LT snapshot), missed events accumulate and the observed event count intermittently exceeds the predicted one. The model, however, responds with intermediate probabilities rather than confident misclassification. A few nodes retain elevated probabilities into the recovery (05:00 LT snapshot), yielding residual false alarms before predictions return cleanly to quiet after 06:00 LT. The contrast between the sharp, well-anticipated onset and the gradual, harder-to-track decay is consistent with the model inputs. For post-sunset irregularities, the history window can capture precursor information through temporal variability in TEC and the local solar time encoding of the solar-terminator transition. After onset, persistence and decay may depend more strongly on local propagation conditions and small-scale irregularity dynamics that are not represented by the model inputs.

\textbf{Event 2 (2025-01-25):} Fig.~\ref{fig_event2} presents a disturbed night during which one receiver of the pair (SIN1) suffered an extended data outage while the companion station (NTUS) maintained 16--25 valid LoS throughout. The night therefore offers a real-world case of coverage loss, affecting both the model input and the testable label set. Before 20:00~LT, only NTUS observes and predictions remain quiet. The two receivers co-observe between 20:00 and 21:20~LT; the event detected in this interval and its segments are therefore cross-confirmed (Section~\ref{sec:qc_labels}). Unlike in Event~1, the model misses the onset entirely. Event~2 combines an abrupt onset with no apparent precursor in the observables and a history window reduced to single-station graphs. Predictions therefore arise only reactively, once the onset itself enters the history window. At a +60\,min lead, this bounds the first possible detection to 60\,min after onset, and the model reaches exactly this bound: missed events accumulate from the onset until roughly 21:20~LT, one lead time later, after which predicted and observed event counts track closely from about 21:30~LT through 03:00~LT (23:00 and 01:00~LT snapshots). When SIN1 drops out again, newly detected segments can no longer be corroborated and are excluded as unverified. The testable set therefore collapses to as few as 3--6 confirmed LoS (the labeled remainder of the segments confirmed at onset). Meanwhile, NTUS continues to sample the disturbance, with mean ROTI fluctuating between 0.5 and 1.2\,TECU/min until about 02:00~LT, keeping the history window informed. As in Event~1, the remaining errors cluster around the decay: elevated probabilities persist into the recovery, producing a false-alarm burst near 05:30--06:30~LT before predictions settle after full two-station coverage returns at 08:00~LT.

Event~2 exposes an evaluation limit: events detected during an outage cannot be cross-confirmed, so forecast skill over the outage period can be scored only on the segments confirmed before it. Quantifying prediction under coverage loss therefore requires a controlled experiment in which every node retains its label, which Section~\ref{sec:coverage_dropout} provides.

\begin{figure*}[!tb]
\centering
\includegraphics[width=\textwidth]{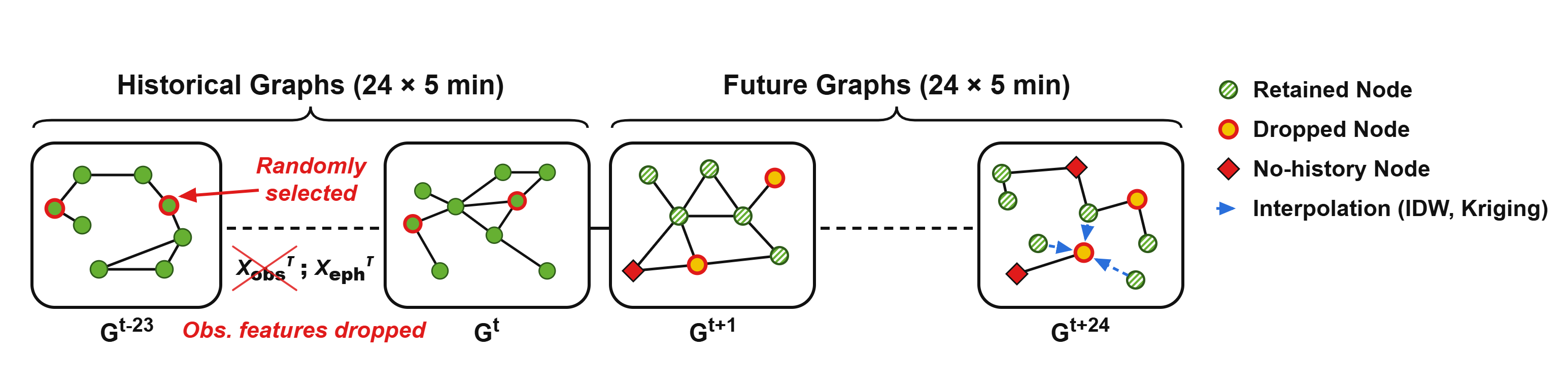}
\caption{Coverage-dropout simulation. In the historical graphs, randomly selected nodes (red outlines) have their observed features $\mathbf{X}^{\tau}_{\mathrm{obs}}$ zeroed and masked out, while their ephemeris-derived features $\mathbf{X}^{\tau}_{\mathrm{eph}}$, positions, and $K$-NN edges are preserved. In the future graphs, retained nodes (hatched green), dropped nodes (yellow), and no-history nodes (red diamonds) are evaluated separately. Blue arrows indicate interpolation (IDW, kriging) of a dropped node from observed neighbors.}
\label{fig_dropout_schematic}
\end{figure*}

\begin{figure*}[!t]
\centering
\includegraphics[width=\textwidth]{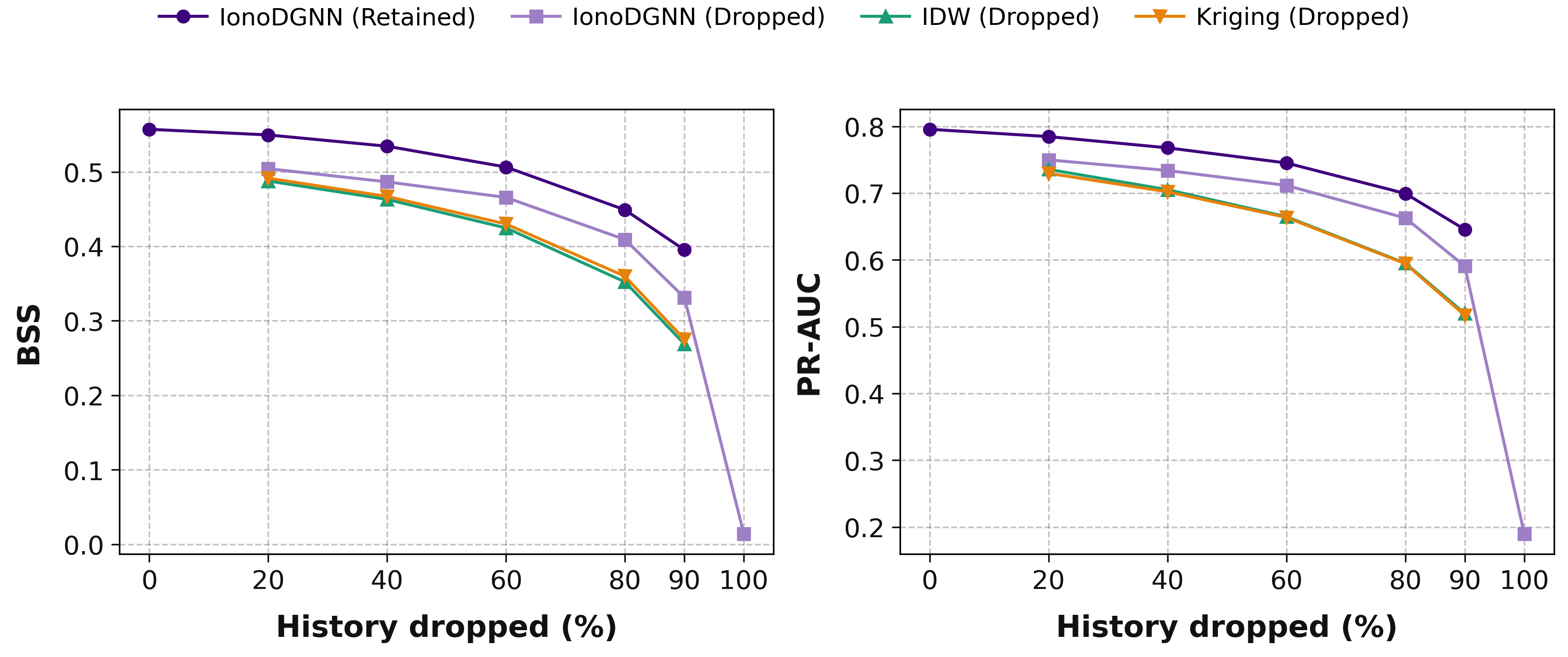}
\caption{BSS and PR-AUC as a function of the fraction of history observations dropped, evaluated separately for retained and dropped nodes, with IDW and kriging interpolation of the dropped nodes as baselines.}
\label{fig_coverage_dropout}
\end{figure*}

\subsection{Prediction Under Coverage Dropout}
\label{sec:coverage_dropout}
Coverage outages and data gaps are common in operational GNSS networks, and Event~2 (Section~\ref{sec:event_vis}) shows that outages during events create missing ground truth. We therefore simulate coverage loss on the fully labeled test set, considering the case where observations become temporarily unavailable in the history window. This is distinct from the no-history scenario (Section~\ref{sec:masks}). In this case, the LoS was observable in the history window but its observations are lost. The node therefore retains its position in the graph topology and its $K$-NN edges, while only its observed features $\mathbf{X}_{\mathrm{obs}}^{\tau}$ (Section~\ref{sec:graph_construction}) become unavailable. To simulate this, we zero $\mathbf{X}_{\mathrm{obs}}^{\tau}$ and set the corresponding entries of $\mathbf{M}_{\mathrm{hist}}$ to zero for a fraction of history-window nodes, preserving the graph structure (Fig.~\ref{fig_dropout_schematic}). In what follows, dropped and retained refer to a node's observations, not to the node itself. Dropped nodes are skipped by the history encoder but remain connected to their neighbors, allowing information to flow through the history module's GNN.

Note that Event~2 differs from this simulation as well: because graph construction instantiates nodes only for observed LoS, the SIN1 outage removed the affected nodes from the history graphs entirely, rather than retaining their positions and edges. The simulation therefore represents a counterfactual in which the outage is instead encoded with placeholder nodes, allowing the model to propagate information from observed neighbors. This convention is attainable in practice, since IPP positions depend only on ephemerides and receiver coordinates and can thus be constructed for a known receiver outage.

As in Section~\ref{sec:nohistory}, dropped nodes are also interpolated from retained neighbors using IDW and kriging as baselines. No-history nodes are excluded from this experiment for simplicity, since they are already evaluated in Section~\ref{sec:nohistory}.

Fig.~\ref{fig_coverage_dropout} shows BSS and PR-AUC as a function of the fraction of history observations dropped. Performance degrades gracefully: retained nodes decline from BSS $\approx$ 0.56 at full coverage to $\approx$ 0.40 at 90\% dropout. Dropped nodes track them closely, still retaining BSS $\approx$ 0.47 and PR-AUC $\approx$ 0.71 at 60\% dropout. At every dropout level, IonoDGNN's dropped-node predictions exceed both interpolation baselines, with the margin widening as coverage thins (BSS 0.33 vs.\ $\approx$0.27 for either interpolator at 90\%). This indicates that learned spatial aggregation through the graph extracts more context from the remaining observed neighbors than distance-based interpolation. The two interpolators nearly coincide across the sweep. Under random dropout, a dropped node's co-located counterpart often remains among the sources, so both exact interpolators recover its value, and at high dropout too few sources remain for their weighting schemes to differ. No-history nodes lack such a counterpart, which is why the two methods separate in Section~\ref{sec:nohistory}. At 100\% dropout, no observed neighbors remain and dropped-node skill collapses (BSS $\approx$ 0, PR-AUC $\approx$ 0.19). This confirms that the recovery is driven by information propagated from observed nodes rather than a learned climatological prior.

Finally, to revisit the station-level loss of Event~2 under this convention, we evaluate a complete single-receiver blackout. Dropping the observations of all SIN1 nodes produces results comparable to 50\% random dropout in Fig.~\ref{fig_coverage_dropout} (BSS 0.44 for dropped nodes, 0.51 for retained). This result is specific to our setup, as the two stations are co-located. In larger and geographically distributed networks, a station outage would instead withhold observations from a spatially coherent cluster of nodes with no immediate observed neighbors. We leave this harder case to future work.

\begin{figure*}[!tb]
\centering
\includegraphics[width=\textwidth]{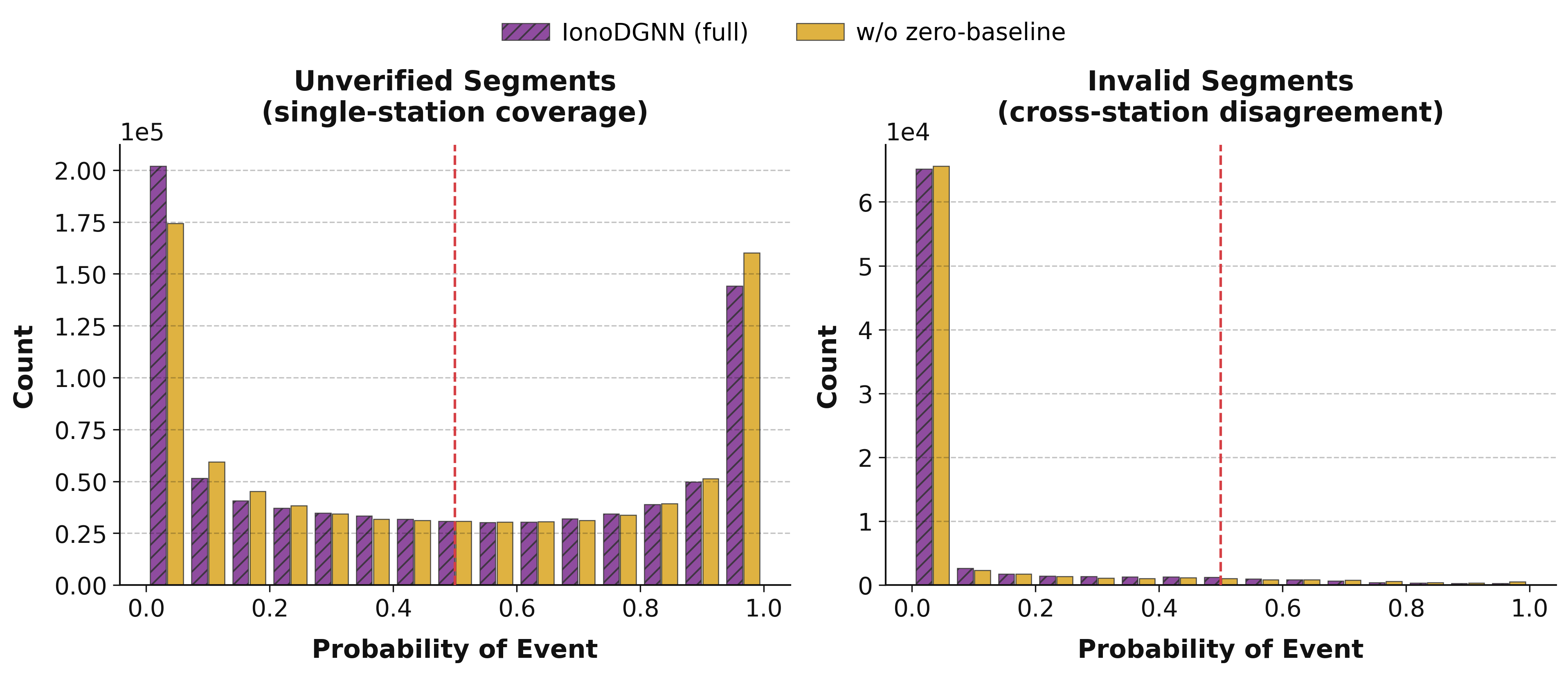}
\caption{Predicted event probabilities for samples excluded from training. (a)~Unverified segments, where the second station lacks data. (b)~Invalid segments, where the second station has valid but quiet data. Each panel shows distributions for the full model and the variant with the zero-baseline edge feature removed.}
\label{fig_ambiguous}
\end{figure*}

\subsection{Sensitivity to Labeling Strategy}
\label{sec:ambiguous}

A potential concern with zero-baseline QC labeling is that the model could learn to associate events with the presence of corroborating data from the second receiver, rather than with spatially coherent ionospheric structure. We test this using the ambiguous labels excluded from training (Section~\ref{sec:qc_labels}). As we use a specific flag for zero-baseline edges, we can also test whether the model relies on this explicit feature or instead learns to infer spatial coherence from the graph neighborhood.

Fig.~\ref{fig_ambiguous} shows the distribution of predicted event probabilities for these held-out samples. Unverified segments, where the second station has no data coverage during the event window, produce a bimodal distribution ($\bar{p} = 0.511$, 52\% exceeding 0.5). If the model relied on cross-station confirmation, these cases would uniformly receive low probabilities. Instead, over half are assigned high event probability, indicating independence from the labeling mechanism. Invalid segments, where both stations have valid data but disagree, receive overwhelmingly low predictions ($\bar{p} = 0.085$, 7.5\% exceeding 0.5). This shows the model can identify data inconsistencies without access to QC labels: when a single IPP reports elevated ROTI but its spatial neighbors remain quiet, it assigns low event probability.

To confirm that this behavior does not depend on the explicit zero-baseline edge feature, we evaluate a variant with this flag removed (w/o zero-baseline). The distributions are nearly identical (Fig.~\ref{fig_ambiguous}), indicating that these predictions are driven by spatial coherence in the graph neighborhood rather than by the explicit station-pairing feature.

\section{Discussion}
\label{sec:discussion}

\subsection{Limitations}
\label{sec:limitations}

The dataset is limited in both spatial and temporal coverage, which constrains the range of events available for training and testing. The setup comprises two co-located stations in a single equatorial region across ${\sim}2$ years of Solar Cycle 25. While this period includes frequent post-sunset activity, no confirmed test events coincide with major geomagnetic storms, leaving storm-driven irregularities untested. Extending the dataset is constrained by the QC scheme, which requires a co-located station pair, and by limited IGS coverage in Southeast Asia. The QC scheme also limits event evaluation during coverage gaps, as shown in Section~\ref{sec:event_vis}: the model produces probabilities for every LoS, but only cross-confirmed segments are testable. Verifying such forecasts in a real-world gap therefore requires a denser station network or an independent reference such as ROTI maps~\cite{Abadi2025ROTI}.

This constraint also leaves single-station performance outside the scope of the present study. However, segmentation heuristics such as rise slope and segment duration (Figure~\ref{fig:fig_qc_examples}) could be used to filter out invalid segments, and such methods can be tested against the cross-confirmation available in the current dataset. Establishing such methods would allow larger dataset with less limitations bound by labeling. Future studies will work towards that to determine whether such a dataset yields similar results.

The 0.3\,TECU/min event threshold used for segmentation is a design choice reflecting the amplitude statistics of the events in this dataset (Section~\ref{sec:qc_labels}). For operational use, it should be tuned to the specific application: lower for early warning, higher to reduce false alarms.

Section~\ref{sec:event_vis} suggests that the model struggles to determine when an event ends. During the decay phase, its predictions appear to be dominated by the persistence of the irregularities rather than by signs of their weakening. One way to address this limitation is to introduce features that explicitly capture the weakening of the irregularities. For plasma-bubble-related events, ionogram traces showing a weakening F-layer signature could provide such an additional feature~\cite{Gon20}.

Although the results demonstrate the feasibility of the change in data representation, the potential benefits of dynamic graphs need further testing. Isolating the contribution of the representation requires direct comparison against equivalent grid-based baselines with independent design choices~\cite{foster2008evaluation}, which we leave to future work.

Lastly, our implementation uses observed rather than predicted orbits and is therefore not yet designed for operational use. However, ultra-rapid products provide predicted precise orbits at ${\sim}5$~cm accuracy in real time~\cite{IGS_IGSUSUM_2000}, which is sufficient for our 2-hour forecast horizon. Longer-term forecasts would nevertheless be bounded by the current IGS ultra-rapid prediction window to at most ${\sim}21$ hours ahead.

\subsection{Broader Implications and Future Directions}
\label{sec:broader_implications}

A key implication of this work concerns the role of data representation in ionospheric forecasting. Operating directly in GNSS observation space lets the model exploit the native LoS sampling structure without requiring gridding or spatial smoothing. Spatial relationships among simultaneous observations are explicitly modeled. The consistent gains of the graph-based variants, together with the behavior on ambiguous labels (Section~\ref{sec:ambiguous}) and the spatial recovery under coverage dropout compared to interpolation (Section~\ref{sec:coverage_dropout}), indicate that these relationships carry meaningful information for irregularity forecasting.

Retaining the native LoS structure also enables conditioning on the known future sampling structure. Through ephemeris conditioning, the model incorporates where and how the ionosphere will be sampled in future steps. This yields a capability specific to GNSS: forecasting on no-history nodes, LoS that enter the observation set only within the horizon and therefore carry no observed past at all (Section~\ref{sec:nohistory}).

The dynamic graph framework opens several directions for further research. The graph topology itself is a largely unexplored design space: the current $K$-NN construction is purely geometric. Edges could instead encode physical priors such as magnetic field line connectivity, reflecting the field-aligned structure of ionospheric irregularities~\cite{Bhattacharyya2022EPB}. Node features could likewise be augmented with additional physical drivers such as vertical plasma drift associated with pre-reversal enhancement~\cite{Abadi2022EPB}. Because historical and future graphs can be constructed separately, such physics-informed modifications remain compatible with the conditioning mechanism. The framework is similarly agnostic to the choice of observable: ROTI is used here, but scintillation indices ($S_4$, $\sigma_\phi$) or other ionospheric parameters could be used instead.

Per-LoS prediction is particularly relevant to safety-critical ground- and satellite-based augmentation systems (GBAS/SBAS), whose availability can be significantly degraded by post-sunset irregularities~\cite{MariniPereira2023GBAS}. For these systems, regional disturbance proxies are not necessarily informative, as high ROTI can occur without producing scintillation, and scintillation can occur without affecting system availability. Because predictions are made per LoS, the framework can instead target operationally relevant quantities, such as per-LoS availability, ionospheric gradients, or dilution of precision. This can provide timely warning ahead of stressed periods and improve operational planning for aviation.

Scaling to larger and more diverse observation networks is an immediate next step. As more stations are added, the graph becomes denser and spans a broader spatial extent, capturing regional evolution at the cost of higher computational load. A more substantive extension is heterogeneous graphs~\cite{Tang2025DAHG} with distinct node and edge types per sensor, allowing GNSS to be combined with other ionospheric sensing systems while letting each retain its native sampling. This requires careful graph design and attention to cross-sensor data consistency, but avoids forcing measurements onto a common grid through imputing or resampling.

The predictions on ambiguous labels (Section~\ref{sec:ambiguous}) suggest that the model implicitly learns aspects of data consistency through spatial coherence. When an anomalous signal appears on a single IPP while its neighbors remain quiet, the model assigns low event probability without explicit quality flags. In a denser graph this property could complement existing quality control by providing a continuous, learned assessment of data reliability alongside the forecast itself. This remains to be tested in larger and more diverse networks.

Finally, ephemeris conditioning as a concept is not specific to ionospheric applications: it relies only on known future sampling locations, a property shared by satellite altimetry, SAR interferometry, and radio occultation. Whether the gains observed here transfer to those domains is an open question.

\section{Conclusion}

To the best of our knowledge, this work represents the first dynamic graph model for ionospheric forecasting that operates directly on GNSS lines of sight. The proposed framework also introduces ephemeris conditioning, which uses satellite orbits to specify the future graphs.

IonoDGNN achieves a BSS of 0.55 and PR-AUC of 0.77, corresponding to 53\% and 58\% relative gains over the persistence baseline. The margin widens with lead time: at the 2-hour horizon, IonoDGNN retains BSS $\approx$ 0.40 while persistence approaches zero skill. Graph message passing contributes a 38\% BSS and 19\% PR-AUC improvement over the no-graph ablation, and ephemeris conditioning contributes 13\% and 10\%, respectively. Conditioning enables prediction on no-history nodes. On these nodes, the model retains most of its skill (BSS 0.53, ROC-AUC 0.96) and improves BSS by 22\% over ordinary kriging and 29\% over IDW interpolation.

Under simulated coverage dropout, the model retains skill on affected nodes by aggregating context from observed neighbors, exceeding both interpolation baselines at every dropout level with the margin widening as coverage thins. On ambiguous labels excluded from training, the model assigns low event probability to single-IPP signals inconsistent with their spatial neighborhood, indicating that spatial coherence can be a quality-control signal learned implicitly from the data.

The framework imposes no constraint on the choice of observable and can in principle be scaled to larger heterogeneous networks and physics-informed graph construction. Demonstrating these extensions is left to future work. The current implementation is not designed as an operational model. Future work will also focus on operational deployment, together with evaluation practices that can score every LoS in the future graphs rather than only the cross-confirmed subset. The dataset, pipeline, and trained models are available at \url{https://github.com/Mert-chan/IonoDGNN}.

\vspace{-6pt}
\section*{Acknowledgments}
This work was supported by Nanyang Technological University (NTU) and Thales Alenia Space. The authors wish to thank Erick Lansard, Yung-Sze Gan, Mathias Vanden Bossche, Sebastien Trilles, and Louis Hecker of Thales Alenia Space for their helpful discussions and support during the course of this work. Claude (Anthropic) was used for programming, language and figure editing support during manuscript preparation. All scientific content, results, and conclusions are the authors' own.

\section*{Data Availability Statement}
The 30-second GNSS observation data for stations NTUS and SIN1 are available through the IGS data archive~\cite{Dow2009IGS}. Precise satellite orbits are available from the Wuhan University ultra-rapid combined products~\cite{Li2022iWUH}. Geomagnetic and solar indices are sourced from the NASA OMNIWeb database~\cite{King2005}. GNSS data were downloaded using GAMPII-GOOD~\cite{zhouforme0318_GAMPII-GOOD}, available at \url{https://github.com/zhouforme0318/GAMPII-GOOD}. The IonoDGNN code, processed dataset, and trained models are available at \url{https://github.com/Mert-chan/IonoDGNN}.

\bibliographystyle{IEEEtran}
\bibliography{ref}

\vfill

\end{document}